# How ChatGPT Changed the Media's Narratives on AI: A Semi-Automated Narrative Analysis Through Frame Semantics

Igor Ryazanov[a], Carl Öhman[b], Johanna Björklund[a]
[a]Department of Computing Science, Umeå University, Umeå, Sweden
[b]Department of Government, Uppsala University, Uppsala, Sweden

## Abstract

We perform a mixed-method frame semantics-based analysis on a dataset of more than 49,000 sentences collected from 5846 news articles that mention AI. The dataset covers the twelve-month period centred around the launch of OpenAI's chatbot ChatGPT and is collected from the most visited open-access English-language news publishers. Our findings indicate that during the six months succeeding the launch, media attention rose tenfold—from already historically high levels. During this period, discourse has become increasingly centred around experts and political leaders, and AI has become more closely associated with dangers and risks. A deeper review of the data also suggests a qualitative shift in the types of threat AI is thought to represent, as well as the anthropomorphic qualities ascribed to it.

## 1. Introduction

Few recent topics have attracted as much attention within popular media as artificial intelligence (AI). This intensive coverage has arguably become an important force in shaping the public's perception of AI, which may ultimately affect both the direction and scope of research, technology adoption and subsequent regulation (Cave et al., 2018; Nader et al., 2022). As such, analysing dominant narratives around AI in popular media is becoming a key priority within the academic literature (Cave et al., 2018).

Despite an abundance of studies in recent years, a number of methodological limitations persist. These often take the form of a trade-off between sample size and depth of analysis (for which we account below), but also of an excessive focus on US and UK 'quality outlets'. More important still, the majority of studies were conducted before the launch of OpenAI's ChatGPT and the subsequent explosion in interest in AI and large language models (LLMs). A brief examination of Google trends data suggests that public interest in AI surged dramatically immediately after the chatbot's launch, reaching levels several orders of magnitude higher than the preceding year (Figure 1). It is plausible that this increase is reflected in media coverage, which would most likely have changed the content and angle of that reporting too. In short, the current literature has not yet taken into consideration what is presumably the biggest event in media reporting on AI in decades.

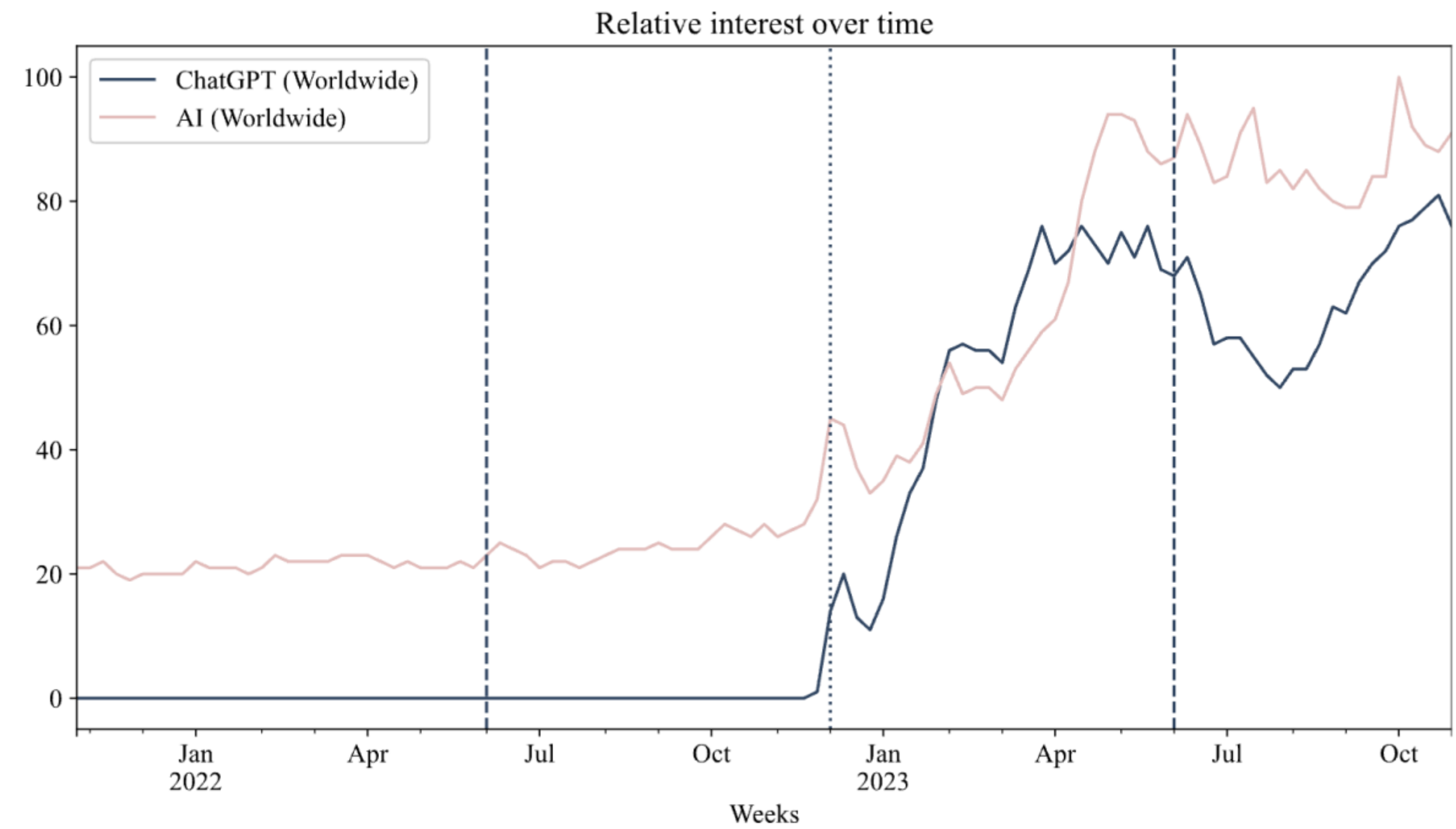

**Fig. 1** Relative interest in 'ChatGPT' and 'AI' between October 2021 and October 2023 based on search queries. Source: Google Trends (https://trends.google.com/trends/, accessed 29 May 2024). Dashed lines indicate the timeframe of the dataset, the dotted line—the launch date of ChatGPT

In view of these limitations, the goal of the present study is to explore how narratives around AI differ before and after the launch of ChatGPT. In pursuing this goal, we draw on a large curated dataset of 5846 articles covering six months before and six months after November 30, 2022 (i.e. the launch date). Rather than targeting a short list of the most reputable sources (New York Times, Washington Post, etc.), we built this dataset from 18 of the most visited anglophone news sites on the open web (as ranked by Majid, 2023). In addition to this new dataset, we also introduce a novel methodology for the study of media narratives based on frame semantics—an analysis of automatically annotated semantic frames in sentences mentioning AI—which, in turn, allows us to identify relevant segments for qualitative analysis. These advantages enable a large-scale yet nuanced analysis of the narratives around AI in the media that are most likely to shape the perceptions of the larger population. We describe our methodological procedure in detail in Section 3, followed by our findings in Section 4, but first, we dedicate Section 2 to a closer review of the literature.

## 2. AI in the media

Existing research on media narratives around AI considers them from several perspectives, and the methodologies are not always directly comparable, leading to somewhat inconclusive results. There are, however, a handful of clearly distinguishable trends. The most prominent is the increase—in some studies by orders of magnitude—in reporting on AI commencing around 2014–2016 (see Brantner and Saurwein, 2021; Chuan et al., 2019; Sun et al., 2020; Zhai et al., 2020). Brantner and Saurwein (2020), for instance, report that about 40% of the mentions of AI from 1991–2018 are found in articles published in the final year of that span.[1] Findings on the sentiments of narratives around AI are also relatively consistent. A majority of studies conclude that media reporting has had an overall positive framing (Brantner and Saurwein, 2021; Brennen et al. 2018; Chuan et al. 2019; Fast and Horvitz, 2017; Frost and Carter, 2020; Köstler and Ossewaarde, 2022), although there are exceptions (Lupton, 2021), and a slightly negative trend over time (Nguyen and Hekman, 2022; Ouchchy et al., 2018). When it comes to topics, it appears that business, healthcare and education are commonly given the positive angles, although it is difficult to make consistent comparisons across studies. The negative and danger-related angles, on the other hand, concern ethical issues, primarily bias and discrimination; job displacement; loss of control; and catastrophic events such as technological singularity and war (see Brantner and Saurwein, 2021; Chuan et al. 2019; Fast and Horvitz, 2017; Zhai et al., 2020).

[1] Nguyen and Hekman (2022) notably find that reporting has actually declined since 2018, proposing that it was the peak year in the 'hype cycle', at least in quality anglophone media.

In addition to these broader trends, one may also note some more specific findings. Brennen et al. (2019) examine which AI experts are granted the authority to shape narratives in the media over 30 years in the US and the UK, noting that the ten most cited AI scholars take up 70% of the media space in both countries. They also find a bias against female experts, who only make up 6% of the sample, and against academically affiliated (as opposed to industry) experts. These findings are echoed in the gender balance of the reporters (about 60/40), although not nearly as starkly (Nguyen and Hekman, 2022). Nguyen and Hekman also find that reporting is dominated by a small group of 'AI alpha journalists' who account for a disproportionate number of articles. For example, journalist Cade Metz authored more than 200 of the articles in their sample of 3098 articles from four US and UK mainstream and tech-related outlets over 11 years.

Another recent topic relating to the reporting and conversations about AI is the degree to which it is being anthropomorphised, such as when an AI system is said to 'understand' something or that someone is 'having a conversation' with the AI. The concerns of such language being misleading and potentially harmful apply to both AI applications—particularly chatbots designed to appear more human-like (Abercrombie et al., 2023)—and to the language of scientific publications and downstream news coverage (Cheng et al., 2024). One such harm is leading the user to place undue confidence in the AI's abilities, which can end in disappointment and distrust. A recent study, based on general public respondents' assessment, shows, however, that the use of anthropomorphising language does not necessarily increase the level of trust people place in technology (Inie, 2024).

In spite of its diversity, the present literature suffers from a number of shared limitations regarding methodology and scope. For instance, with very few exceptions, conclusions about media reporting are based on 'quality print media'; in six of the most comprehensive studies, the New York Times is the dominant source of data (Chuan et al., 2019; Fast and Horvitz, 2017; Nguyen and Hekman, 2022; Sun et al., 2020; Zhai et al., 2020). While an important driver of public discourse, the broader public tends to obtain their news from open-web sources rather than paywalled quality outlets. The focus on quality outlets is also connected to a geographical bias in the data. Although there are notable exceptions from Austria (Brantner & Saurwein, 2021), Germany (Köstler and Ossewaarde, 2022) and Australia (Lupton, 2021), among other regions, the vast majority of studies focus exclusively on American or British news sources.

Regarding methods, there appears to be a trade-off between size and nuance. Studies involving greater samples are typically restricted to identifying broader trends and cannot uncover the discursive patterns that fall beyond topic modelling or sentiment analysis. For example, Nguyen and Hekman (2022) employ unsupervised clustering based on term frequencies and treat the resulting topics as discourse frames. They also perform dictionary-based sentiment analysis and assess the presence of different AI-related risks. The authors discover frames (i.e. topic clusters) in the AI discourse as well as the most common associated risks but do not go beyond word frequencies and do not combine the findings with any further qualitative investigations. This leads to frames being largely topical: 'AI & Governance', 'AI & Healthcare', etc. Studies involving larger samples are also limited in that they typically draw on data collected over several decades of reporting, meaning that general conclusions about 'the state of media reporting' are coloured by outdated materials. Moreover, studies that are based on sentiment analysis, such as Garvey and Maskal (2020), often rely on black-box methods, which makes it hard to interpret the measured metrics or analyse results on a more granular level.

Meanwhile, studies aimed at finding nuance and detail 'between the lines' often draw on small samples —Köstler and Ossewaarde (2022), Frost and Carter (2020), and Lupton (2021) fall into this category with 40–136 articles—limiting their generalisability beyond a specific region or context. There are exceptions, such as Zhai et al. (2020), who use the largest sample of articles (n=9914) yet still conduct a thorough examination using topic modelling, sentiment analysis and co-occurrence analysis for countries and individuals. This kind of study is limited to the correlation of occurrences of key terms, so it can only analyse the dataset on the document level and is not suitable for investigating how the key terms themselves are used. It also may struggle to connect generalised results with the qualitative changes they reflect. Larger-scale qualitative analyses do exist but make other methodological trade-offs: for example, Roe and Perkins (2023) analyse 671 news items but limit themselves to headlines of the UK coverage.

The final limitation of the literature is that the majority of the present studies were conducted before the launch of OpenAI's ChatGPT. As we have already suggested, there are compelling reasons to believe that this event brought major changes to the way the media discusses AI. Yet, to our knowledge, the only available study on this so far is Roe and Perkins's (2023) thematic analysis of headlines that mention ChatGPT and AI in UK media between January and May of 2023. While relevant, this study is subject to the methodological limitations listed above, and focuses only on broader topics and narratives, rather than on the concept of AI itself. As such, ChatCPT's effect on the media's narratives on AI remains largely unexplored.

In summary, while current literature has successfully explored the broader long-term narrative trends within anglophone quality outlets, it struggles to balance scope and depth. Moreover, we do not know how what appears to be a most critical event in AI news reporting—the public launch of ChatGPT—changed these narratives. In the following section, we propose a novel methodology, as well as a unique dataset, to address these limitations.

# 3. Method

Below, we present the data and analytical procedure used in our study and comment on methodological limitations.

## *3.1. Data*

This article aims to capture the narratives around AI encountered by the broader public. This is in contrast to focusing only on 'quality outlets', which may drive expert discourse but have a more limited reach (Chuan et al., 2019; Fast and Horvitz, 2017; Nguyen and Hekman, 2022; Sun et al., 2020; Zhai et al., 2020). To this end, we used Press Gazette's list of the world's top online news sources, ranked by traffic according to Similarweb data (see Majid, 2023).[2] From the initial list of 25 sites, we removed those that require a subscription due to the technical difficulty of retrieving content from paywalled sites, not to mention the ethical and legal issues involved. This led to, e.g. the New York Times and Washington Post being taken off the list. We note that MSN do not have their own news editors and aggregates news from other sources, so although the site requires the removal of duplicate content, it also provides additional coverage outside of our list of named sites. The resulting list contains the 18 outlets reported in Table 1.

**Table 1.** 18 news websites from which the dataset was collected.

| | | |
|---|---|---|
| BBC<br>bbc.com | Fox News<br>foxnews.com | New York Post<br>nypost.com |
| BuzzFeed<br>buzzfeed.com | The Guardian<br>theguardian.com | News18<br>news18.com |
| CNN<br>cnn.com | Hindustan Times<br>hindustantimes.com | People<br>people.com |
| Daily Mail<br>dailymail.co.uk | India.com<br>india.com | The Sun<br>the-sun.com |
| Daily Mirror<br>mirror.co.uk | MSN<br>msn.com | The Times of India<br>timesofindia.indiatimes.com |
| Forbes<br>forbes.com | NDTV<br>ndtv.com | USA Today<br>usatoday.com |

[2] The ranking from June 2023, during the data collection phase of the presented work.

Data collection was conducted through a commercial data retrieval service and directed by Google Search. The search queries followed the format: ‘<keyword> site:<source site>’, and the results were filtered to limit hits to content published from the 31st of May 2022 until the 31st of May 2023. Three keywords—*AI*, *ChatGPT* and *Machine Learning*[3]—were used for each of the 18 sites, amounting to 54 searches. For each news article web page, full text and metadata were retrieved. The metadata includes headlines, preambles (when applicable), and publication timestamps but excludes authors, links to non-textual media, and HTML markdown. At the completion of this procedure, a total of 15,539 articles were retrieved. Due to using several related keywords, as well as the general practice of republishing the same articles with minor changes (e.g. under a new headline), and the search engine not being entirely reliable in distinguishing the main content of the webpage, the dataset required filtering. The cleaning process involved removing 132 articles written in languages other than English and 4827 articles that were complete or near-complete duplicates. Finally, 4525 articles did not explicitly mention AI in their body (but rather in, e.g. the page markdown), and these were also omitted. This left a total of 6055 articles for the working dataset, out of which 5846 mentioned AI and 209 mentioned ChatGPT without bringing up AI. In the final step, we identified 49,120 sentences talking explicitly about AI. These sentences form the basis of our analysis.

### *3.2. Analytical procedure*

Rather than conducting formal hypothesis testing, this study takes an exploratory approach, which allows us to examine interesting patterns in the data as they emerge. While predefined hypotheses are normally preferable within quantitative social science, the present study does not seek to test the validity of any particular scientific theory but to characterise an arguably unique event where expectations are not clearly defined. This motivates our choice of a more open-ended, probing approach.

Below is a summary of our methodological procedure, followed by a detailed explanation of each step:

1. Calculate and analyse the general statistics of the articles in our dataset;
2. Analyse frame annotations in general:
    a. Annotate the sentences mentioning AI or ChatGPT with semantic frames;
    b. Analyse and interpret the most significant relative changes since the release of ChatGPT;
3. Analyse frames mentioning AI:
    a. Calculate the number of frames that include AI as one of the frame elements and compare their frequencies before and after the release of ChatGPT;
    b. Analyse and interpret the most significant relative changes;
4. For two topics of interest—the danger of AI and AI anthropomorphism—perform the following analysis:
    a. Manually curate a list of relevant frames and/or frame elements;
    b. Analyse and interpret the most significant relative changes since the release of ChatGPT;
    c. Use the list of frames to subsample 200 topical sentences from before and after the release of ChatGPT;
    d. Annotate the sentences according to the themes of the topic and analyse the breakdown. The annotation is done by the two independent annotators labelled as A1 and A2.

As the first step of our analysis, we look at the frequencies with which the term ‘AI’ is mentioned in the news coverage. We compare the coverage of different publications in our dataset and correlate the coverage of ChatGPT and AI in general (Section 4.1). Having considered the high-level statistics of references to AI, we delve deeper into specific uses of the term. For this purpose, we apply frame semantics, a descriptive framework that characterises lexical units in terms of semantic frames. A semantic frame is here defined as ‘a schematic representation of a situation involving various participants, props and other conceptual roles, each of which is a frame element’. Frame elements (FE) are frame-specific semantic roles that can be seen as additional arguments of the frame (e.g. a *Statement* frame is likely to have a *Speaker* frame element). In our annotations and throughout this article, we use the frame and FE names from FrameNet, a lexical database annotated according to the frame semantics theory (Baker et al., 1998).

[3] *Machine Learning* was used in the data collection, but analysis of the dataset showed its redundancy as a keyword.

By analysing the types of semantic frames invoked when AI is mentioned, as well as AI's role in these sentences (i.e. which element of the frame it realises), we aim to map the main components of the AI-related discourse. Annotating the texts with semantic frames also allows manual analysis of specific usages of the term by aggregating the relevant mentions. Manual analysis and annotation of thousands of articles is impractical, so we apply a specialised information-extraction model to annotate the sentences automatically using LOME (Xia et al., 2021). LOME is a modular system for entity typing and semantic role labelling. At its core is a parser trained on FrameNet (Baker et al., 1998). For our task, we use this core functionality to annotate the sentences mentioning AI. An example annotation of a single semantic frame is presented below. The frame *Warning* is invoked by the lexical unit 'warning' and has two additional frame elements: *Speaker* and *Topic*.

| Why are artists warning against using AI-generated images? | | |
|---|---|---|
| Frame | *Warning* | 'warning' |
| Frame elements | *Speaker* | 'artists' |
| | *Topic* | 'against using AI-generated images?' |

To determine how the use of the term AI has changed over time, we compare the frequencies with which the frames occur in sentences about AI. We look at them to describe the types of context that surround AI in general. To analyse the use of the term itself, we consider the frames that either have AI as (i) a lexical unit invoking them or (ii) a named frame element. This type of analysis focuses on the situations where AI is invoked and can help trace the changes in the ways the term itself is used. As such, it is different from sentiment analysis, which is common in this type of computational research but considers only the emotional valence of the language without any other context. For frames and frame-FE combinations that display the biggest changes in frequencies, we investigate the annotated sentences to provide an interpretation. In addition, frame-semantic annotations provide a relatively easy way of filtering the uses of the term. Unlike, for example, the output topic modelling, supervised annotations do not have to be individually investigated to infer their meaning.

Taking advantage of the innate meaning of frames, we curate a list of specific frames and frame elements to conduct a deeper analysis of two particular subthreads of AI discourse prominently covered in literature: AI as a source of danger and anthropomorphism of the technology. Besides considering the frequencies or occurrences for these two scenarios, we use curated lists of frames and frame elements to sample relevant sentences and perform two-level thematic analysis. This way, we investigate not only how present danger and anthropomorphism are in the discourse but also what forms they take.

### *3.3. Methodological limitations*

Our methodology is an attempt to find a balance between analysing a large corpus of data while also being able to narrow in and qualitatively analyse confined subtopics. To limit parsing time and the volume of annotations generated, we focus only on sentences specifically mentioning AI, thus omitting any AI-related frames that would only be discoverable given a larger context. For this reason, we disregard any implicit references to AI (e.g. pronouns) that could be identified with a coreference resolution model. While we observe and describe changes in the AI narratives, we do not establish formal causality between, for example, the release of ChatGPT and any language changes. Moreover, it is important to highlight that while changes in language likely indicate shifts in discourse, they may not directly map onto each other.

The structure of our dataset also imposes limitations on our analysis. First and foremost, the dataset contains only texts from anglophone media (8 based in the US, 5 in the UK and 5 in India), which means any findings would only extend to English. We intentionally focus on the publishers with the largest coverage, arguing that they are more likely to inform the general public. It can, however, be argued that more specialised (e.g. Wired) and reputable but paywalled (e.g. The New York Times) media are as or more influential in shaping high-level AI discourse. It should also be noted that Google searches are not a completely unbiased way of retrieving media content. For example, hits appear to vary slightly depending on geographical region. All searches in our sample were made from the United States, so we cannot guarantee a complete alignment with searches performed in other regions. There are also indications that Google puts more weight on more recent search results, but it is difficult to ascertain whether this affects the sample. All in all, our assessment is that, while worth considering, these possible sources of bias do not materially compromise the overall integrity of the dataset.

# 4. Findings

## *4.1 General Statistics*

The distribution of the collected articles over time follows the same pattern as the data on Google searches (see Figure 1 and Figure 2a), with a clear increase in the number of publications around the end of November 2022 (Figure 2a). The total number of publications after the ChatGPT release is more than 6 times higher than in the six months before, with the monthly average rising from 132 before to 842 after.

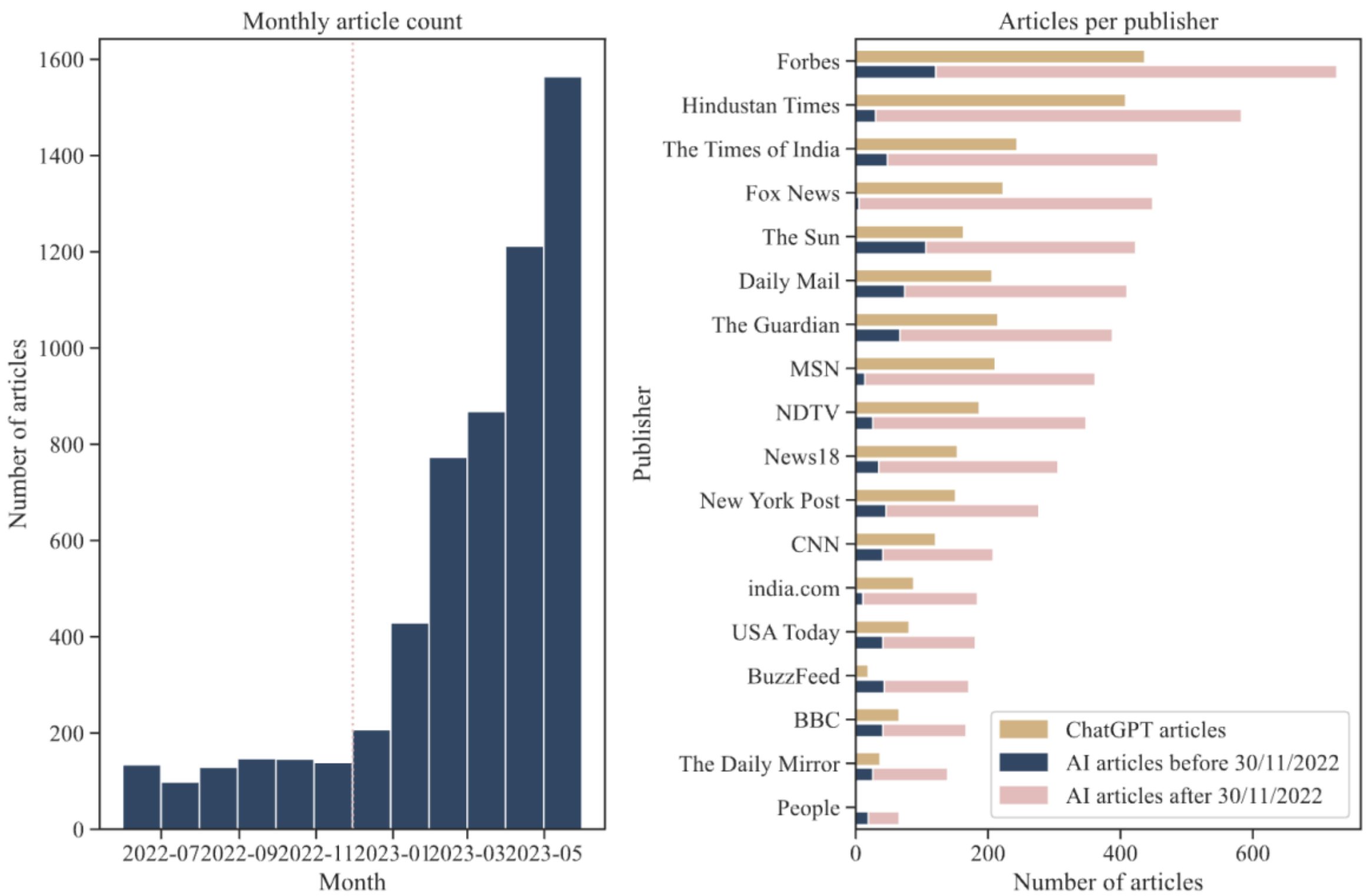

**Fig. 2** The general statistics of the dataset. (2a) Distribution of articles in the dataset over time. (2b) The number of articles per publisher: all articles split by the threshold date of the ChatGPT release, and the articles mentioning ChatGPT specifically[4]

The general increase in coverage is evident, but it is not uniformly distributed across the websites in the dataset. As visualised in Figure 2b, some sites, including Forbes, Hindustan Times and The Times of India, produce far more articles on AI than e.g. People Magazine or The Daily Mirror. We leave it as an open question as to why this is the case. It is also notable that some of the publications only started to provide significant coverage of AI after the release of ChatGPT. In particular, only approximately 1% of Fox News and 3.5% of MSN coverage in our dataset belong to the first six months. MSN does not have its own coverage and serves as an aggregator, so a very low number of AI-related news before the ChatGPT launch suggests this might also be the case for sources outside of our dataset. Similarly, for all 5 of the India-based websites, the coverage in the first half of the dataset was minimal and grew by one order of magnitude after November 2022. The remaining sites have a more substantial amount of AI reporting pre-ChatGPT. On Buzzfeed, for example, the post-ChatGPT coverage is ‘only’ about 3 times higher than before the launch. Forbes was the site that had the largest coverage of AI pre-ChatGPT. Still, it grew more than 6 times post-ChatGPT.

[4] The 209 articles mentioning only ChatGPT but not AI are also included here.

As for the mentions of ChatGPT itself, they are present in 50%–75% of the articles post-ChatGPT for all outlets, with the exceptions of The Daily Mirror (35%), BuzzFeed (16%) and People.com (the smallest sample of 66 articles with only one mentioning ChatGPT). Additionally, 127 articles mention ChatGPT without mentioning AI at all. Out of these, 23 were from MSN, while every other website had less than 15. This suggests that, while much of the interest in AI in the succeeding period can be directly linked to ChatGPT, the release appears to have also sparked an interest in AI more generally. We also explored whether the term ‘AI’ had a higher frequency per article in the succeeding half of the dataset. We found that the mean number of mentions per article, including headlines and preambles, went up by 33% from 8.8 to 11.8 ($p<0.01$), which may be a soft indicator that AI is more often the central topic of the coverage.

### *4.2 Frame analysis*

Filtering the 5846 articles in the dataset resulted in 49,120 sentences mentioning AI. These sentences were passed to the LOME model, which produced 420,155 individual semantic frame annotations. The most common frame, *Statement*, has been annotated 14,246 times. Each frame is evoked by a word or another lexical unit and typically has one or more other frame elements or arguments. Figure 3a illustrates the number of occurrences for the 25 most common frames in the processed sentences. The processed sentences talk explicitly about AI, so aggregated frame annotations can be interpreted as a high-level description of the language surrounding the topic.

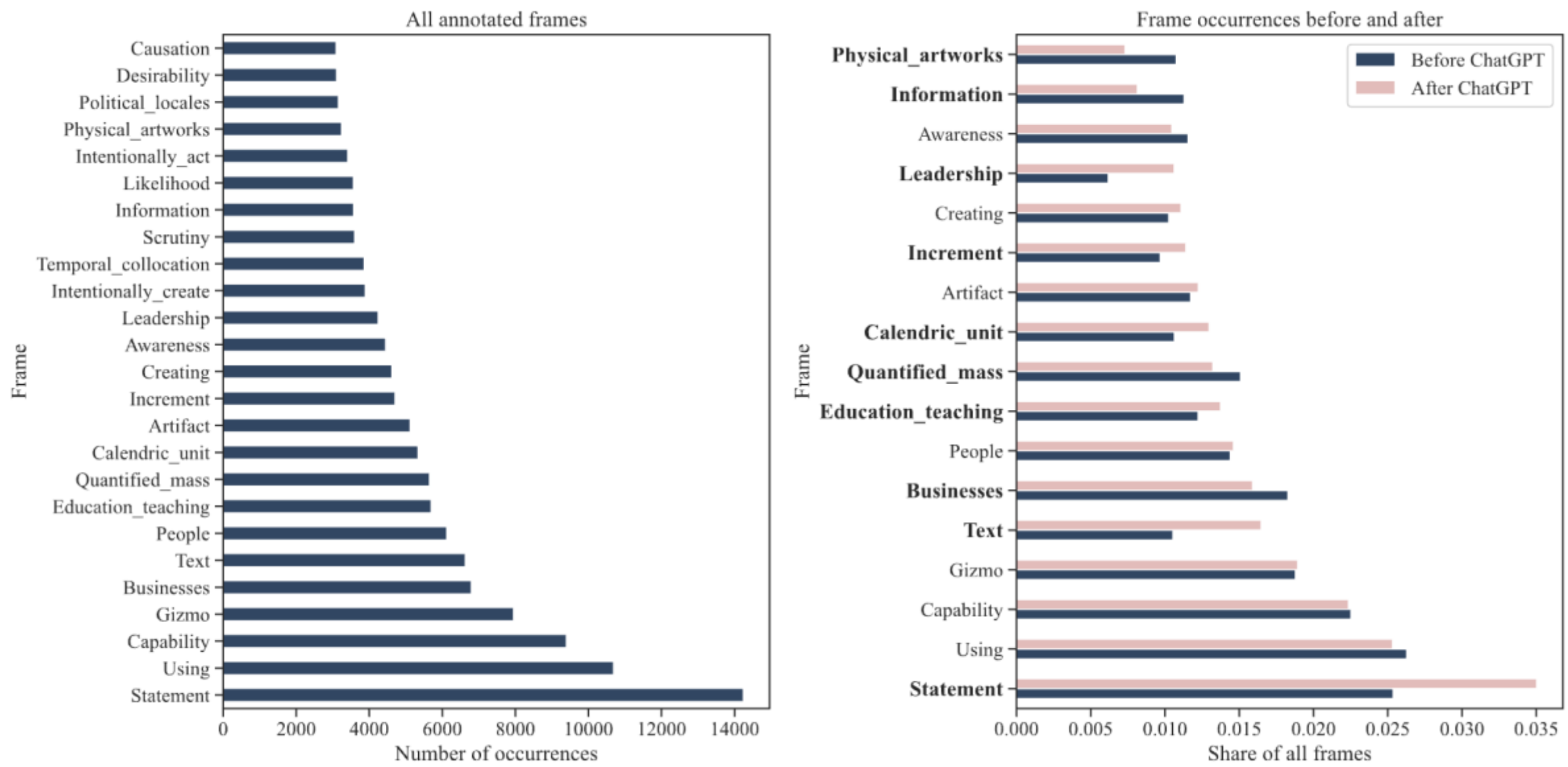

**Fig. 3** The distribution of the most common frames in sentences mentioning AI: to the left is the entire dataset (3a), to the right it is split into groups before and after the release of ChatGPT (3b). 3b shows the combination of 15 most common frames before and after (17 in total). Statistically significant changes ($p<0.01$) in bold

When comparing frame distributions from before and after the release of ChatGPT, even at the frame level, we can see several common frames becoming relatively more or less frequent (Figure 3b). Out of 17 frequency changes showcased in Figure 3b, 10 appear statistically significant at the level of $p<0.01$ (see Online Appendix 1) and support shifts in the discourse around AI. Below, we take a more detailed look at the 4 frames with the biggest relative changes: *Statement*, *Text*, *Leadership* and *Physical_artworks*. For each of them, we provide a definition, an example, and a possible interpretation based on a qualitative review of a subsample of frame occurrences (50 instances per frame).

*Statement*—'This frame contains verbs and nouns that communicate the act of a *Speaker* to address a *Message* to some *Addressee* using language.'[5] Naturally common in news coverage and invoked by quotes and citations, it can reasonably be argued that with the growing public interest, the growth of *Statement* is due to news articles referring more often to experts, press releases (see example below), and in general using more comments by relevant speakers. In addition, AI itself, and especially ChatGPT, assumes the role of the speaker (see also Section 4.3.2) much more often in the second half of our time frame when chatbots suddenly became ubiquitous.

| The New York Times reported Google CEO Sundar Pichai issued a "code red" in response to the threat ChatGPT poses to Google's search business. | | |
|---|---|---|
| Frame | *Statement* | 'reported' |
| Frame elements | *Speaker* | 'The New York Times' |
| | *Message* | 'Google CEO Sundar Pichai issued a "code red" in response to the threat ChatGPT poses to Google's search business.' |

*Text*—'A *Text* is an entity that contains linguistic, symbolic information on a topic, created by an *Author* at the *Time_of_creation*.' This frame is also common in news coverage when referring to papers, articles, letters, posts, etc. Besides increasing coverage and cross-referencing, we expect the relative increase to come from ChatGPT (and analogous systems) introducing the generation of plausible AI texts to the broad audience thus prompting the conversation on 'AI fiction', 'AI poetry', using it as a writing tool, etc.

| ChatGPT is used to write research papers, books, news articles, emails and more. | | |
|---|---|---|
| Frame | *Text* | 'papers', 'books', 'articles', 'emails' |
| Frame elements | *Genre* | 'research' [papers], 'news' [articles] |

*Leadership*—'The frame contains both nouns referring to a title or position, and verbs describing the action of leadership.' In the context of our dataset, the frame refers largely to the people in positions of power: executives, country leaders, and lawmakers. Increasing coverage of such figures in the AI context may be evidence of the increased presence of AI in the domains of business and politics.

| San Francisco: An influential Silicon Valley presence for more than a decade, OpenAI CEO Sam Altman is emerging as the tech titan of the AI age, riding the wave of ChatGPT, the bot his company unleashed on the world. | | |
|---|---|---|
| Frame | *Leadership* | 'CEO' |
| Frame elements | *Leader* | 'Sam Altman' |
| | *Governed* | 'OpenAI' |

*Physical_artworks*—'A physical object, the *Artifact*, is produced by a *Creator* to stimulate the perceptions, emotions, or cognition of an audience' is the frame for which we observe the biggest and most significant relative *decrease* in proportion. Most of the annotations in our dataset correspond to digital images rather than physical artworks, however. It is plausible that this decrease in relative frequency is due to our chosen timeframe. AI-generated images entered public discourse earlier than AI-generated texts (e.g. both DALL-E 2 and MidJourney entered beta in July 2022), so the former's coverage was already growing in the first half of our timeframe.

[5] This and the following frame definitions are taken from https://framenet.icsi.berkeley.edu/ (Baker et al., 1998).

| | | |
|---|---|---|
| Engineers at the Delaware-based company created a "digital twin" of Willis, putting images of his face onto its AI platform, where the engineers are able to create film projects in a matter of days. | | |
| Frame | Physical_artwork | 'images', 'film' |
| Frame elements | Representative | [images] 'of his face' |

To summarise, the changes in frame frequency suggest that the launch of ChatGPT drove public discourse on AI in a direction where leaders and experts play a bigger role than before, as illustrated by the increase of the Statement and Leadership frames. The post-ChatGPT discourse is also clearly more focused on text, plausibly due to the sudden leap in the quality of AI-generated texts, and less on other forms of AI-generated content, such as images. The breakthrough image generation models (e.g. Stable Diffusion) have already been introduced to the public but not yet incorporated into chatbot-like interfaces (e.g. ChatGPT adding DALL·E 3 to its functions in late 2023).

### *4.3 AI as a frame element*

Investigating the frames in the dataset of sentences mentioning AI or ChatGPT gives an overview of the general discourse, but the annotations allow us to go further. By filtering out the frames where the term AI is one of the frame elements, we can get a clearer idea of the roles it plays and the potential changes that may have occurred with the sudden popularity of LLM-based chatbots. To account for the annotations that often include articles and prepositions but avoid the longer phrases that include AI, we filter the strings using a simple regular expression pattern `(^.{0,3} |^)(AI).{0,3}$`. The pattern allows for an up-to-3-letter word before AI and up to 3 characters after. This means that e.g. 'the AIs' would still be counted, but 'OpenAI' or 'an AI artist' would not. While not perfect, this filtering ensures that the vast majority of frame elements refer directly to AI. We save and count frame-frame element combinations, such as (*Assistance, Helper*).

Perhaps expectedly, the most common pairs in both parts of the dataset are from the technology-related frames, such as *Artifact*, *Using* and *Gizmo*. They are followed by a more diverse list of pairs, including AI as a research topic, AI as an agent and AI as a creator. Similar to the analysis in the previous section, we compare the number of occurrences of these pairs before and after the release of ChatGPT (Figure 4b), which does not reveal many significant changes. Among the common frame-FE pairs, only three—(*Gizmo, Use*), (*Project, Salient_entity*) and (*Objective_influence, Influencing_entity*)—have changed on a statistically significant level (see Online Appendix 1). Upon investigation, many occurrences of the (*Project, Salient_entity*) frame pair turned out to be a case of mislabelling of 'AI [computer] programs' as 'AI [research] programs'. (*Gizmo, Use*) simply refers to various 'AI tools' and 'AI systems', the increased frequency of which does not, perhaps, have considerable discursive implications. Finally, the increased frequency of (*Objective_influence, Influencing_entity*) clearly corresponds to the attention to the impact of AI.

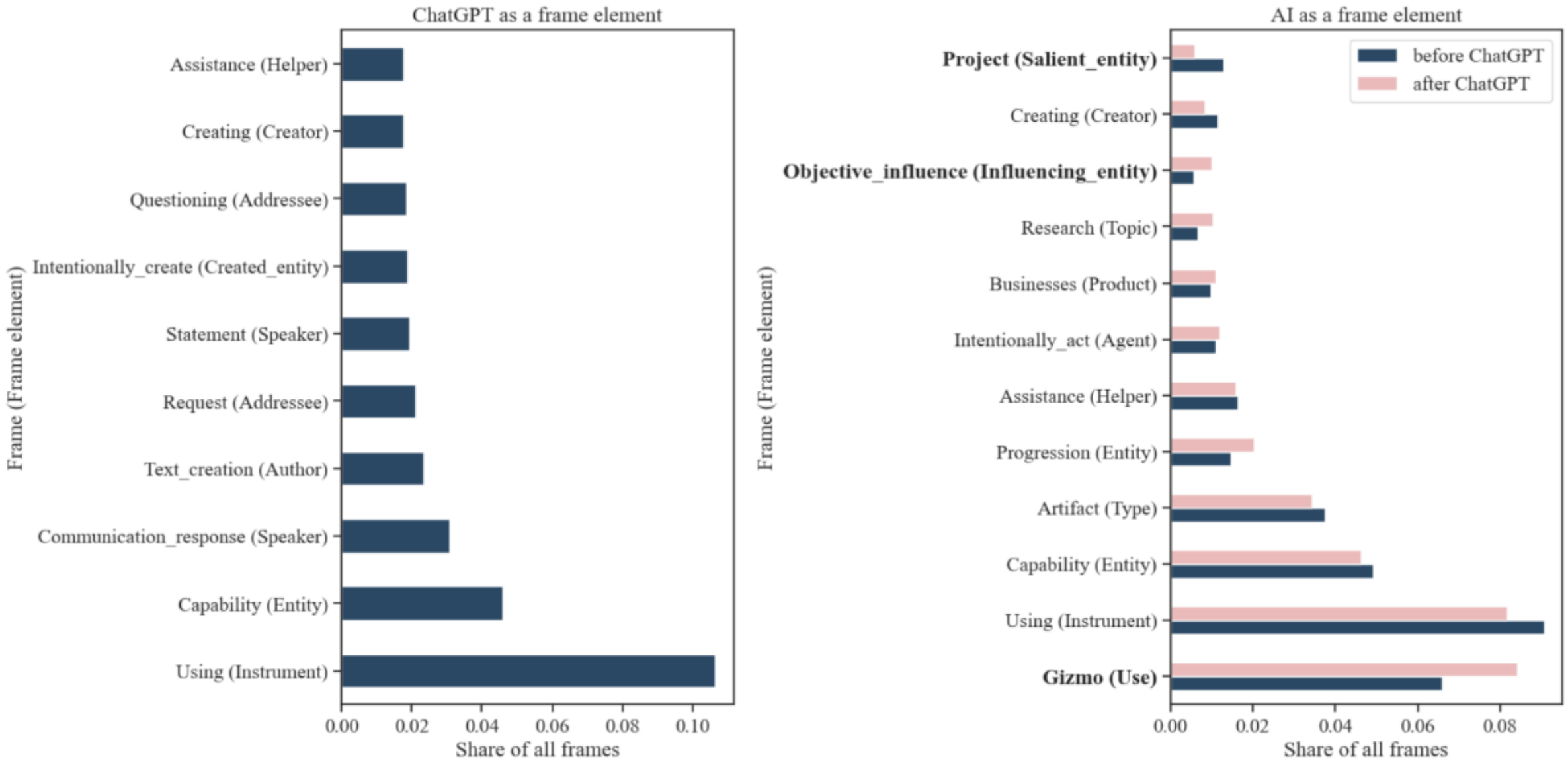

**Fig. 4** To the left are the most common frame-FE pairs that include ChatGPT as a frame element[6] (4a). To the right are the most common frame-FE pairs that include AI as a frame element before and after the release of ChatGPT (4b). Statistically significant changes ($p<0.01$) in bold

A more illuminating comparison is that between AI and ChatGPT. When extracting the frame-FE pairs, where ChatGPT is the frame element, and considering the most frequent ones, we can see only 3 of the top 10 overlap with the ones most common for AI: (*Assistance, Helper*), (*Capability, Entity*) and (*Using, Instrument*) (Figure 4). The remainder are mostly associated with communication and would not be immediately associated with a tool, suggesting that, at least within our dataset, ChatGPT – also representing other chatbots – did not become entirely synonymous with AI upon arrival and, more importantly, that there is a visible tendency of personification or at least communicative intent (e.g. being an *Addressee* of a *Request* or a speaker in *Communication_response*).

The main advantage of considering frame annotations lies, however, in providing a relatively easy and reliable way of automatically identifying instances of much more specific and narrow uses. In the following two subsections, we take advantage of this to investigate AI from two angles commonly considered in the literature: (1) the risks and dangers it relates to and represents and (2) the anthropomorphism of the technology.

### *4.3.1 AI and danger*

As indicated by previous studies, a major theme in the reporting on AI pertains to the potential threats and risks associated with the new technology (see Brantner and Saurwein, 2021; Chuan et al., 2019; Fast and Horvitz, 2017; Zhai et al., 2020). In view of this, we manually curate a set of frames related to risk and danger, as well as to adjacent topics such as deception, weapons and crime. To have sufficient representation of each frame and to make manual selection feasible, we limit the collection to frames that occur at least 100 times. From this list, we manually select those that relate to danger or risk, consulting randomly selected samples of each frame, as well as the FrameNet definitions.

Overall, the share of articles where at least one of the listed frames occurred increased from 37.7% to 44.8% of all articles, indicating that the danger aspect appears to be considered far more commonly in the post-ChatGPT period. Moreover, the total share of the danger-related frames increased from 1.76% in the pre-ChatGPT period to 2.19% in the post-ChatGPT period. At the scale of our sample, this also appears to be a relevant change ($p<0.001$) and indicates that the coverage of AI has become more risk-focused and/or alarmist. To explore what is behind this shift, we look at the relative frequencies of the danger-related frames individually (see Figure 5).

[6] The 209 articles mentioning only ChatGPT but not AI are also included here.

Four frames stand out with the highest relative change: *Predicament* (share down to 0.61 of before), *Warning* (share up to 3 times from before), *Being_at_risk* (share up 1.39 times) and *Risky_situation* (share up to 1.87 times).

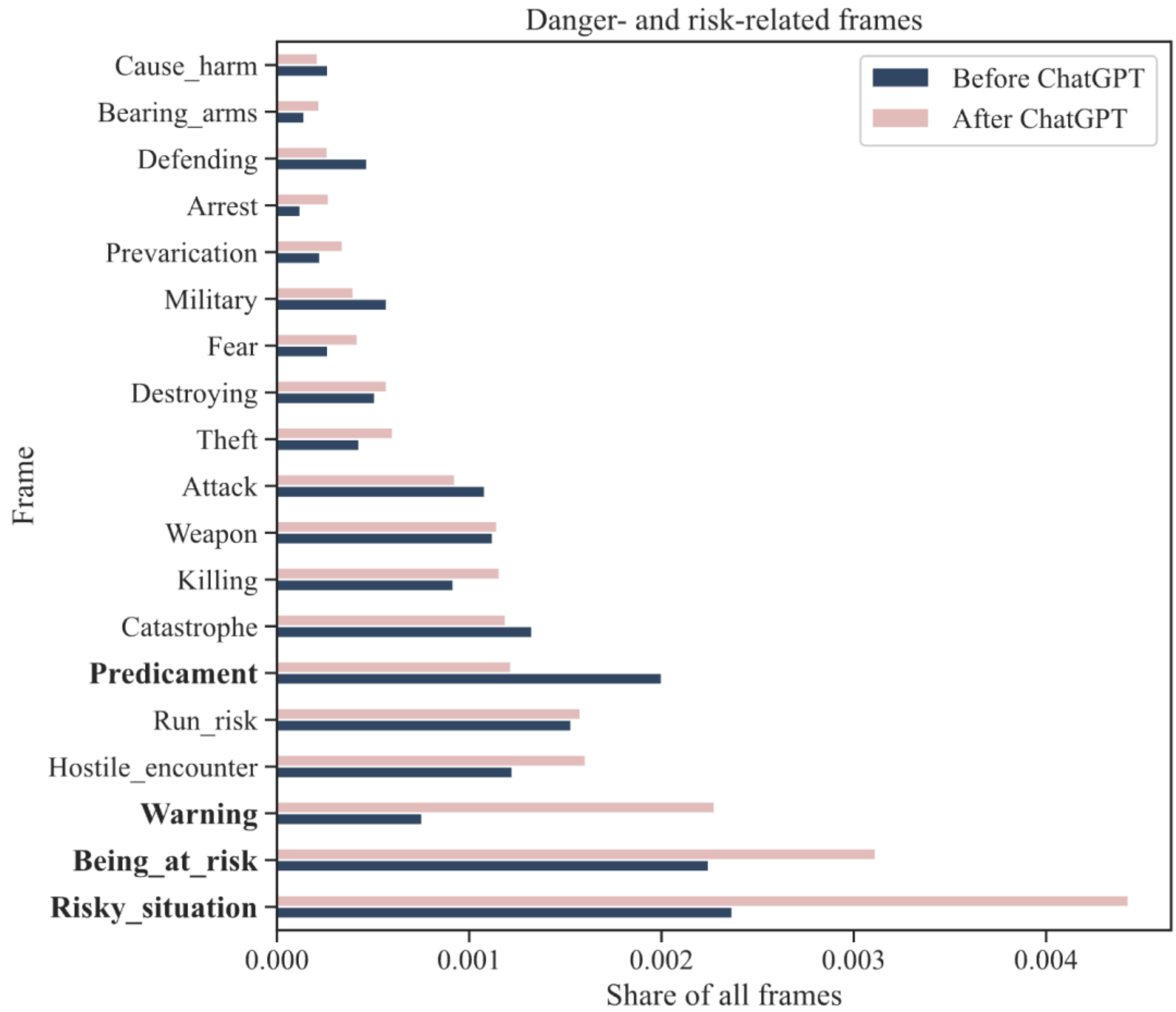

**Fig. 5** Frames related to danger before and after the release of ChatGPT. Statistically significant changes (p<0.01) in bold

The frame *Predicament*, defined as 'An *Experiencer* is in an undesirable *Situation*...' is the only danger-related frame with a significantly decreased share. Essentially, this frame accounts for problems broadly defined. It is triggered most often by the word 'problem' (82% of all occurrences) and rarely includes AI explicitly as an argument. Only on 15 occasions is AI itself an argument, and on 82 occasions it is part of a longer FE string (just 13% of all predicament strings). Even in a set of sentences about AI, the frames' content remains relatively unspecific topically, which may be the reason why it did not show proportionate growth. We fail to ascertain why the frequency of this frame has decreased and leave a deeper analysis as an item for future work.

By contrast, the frame *Warning* occurs much more often during the second half of our timeframe. Similarly to *Predicament*, it is triggered almost exclusively by various verb and noun forms based on 'to warn', and only on 7 occasions 'AI' is an argument. The big difference, however, is that AI is part of at least one argument string in almost 82% of *Warning* frames. In 47% of the cases, AI is mentioned in the *Message* FE, in 30% in *Topic* and as part of *Speaker* in 18%. The much larger share of the *Warning* frame suggests that the coverage of AI shifted significantly towards caution and perhaps even alarmism.

Finally, the frames *Risky_situation* and *Being_at_risk* are the ones that best correspond to the general notion of danger. They are triggered by a diverse set of terms, such as 'risk', 'harm', 'safe', etc. *Risky_situation* focuses on a situation or dangerous entity that may cause harm and does not have to specify what is at risk. Approximately 48% of the *Risky_situation* instances mention AI in their arguments. *Being_at_risk* is the second most common frame and has similar frequency in the first period. It is symmetric to *Risky_situation:* triggering *Being_at_risk* requires the entity at risk being mentioned explicitly (something is in unspecified danger, while *Risky_situation* requires explicit mention of danger). The number of occurrences of *Risky_situation* grew more than *Being_at_risk*, which may indicate not just the increased sense of danger in the AI coverage but also that more risks in the discussion are becoming undirected. This is in line with our following qualitative assessment that the references to 'AI risks' and 'AI dangers' without any additional clarification have become more frequent in the post-ChatGPT coverage.

To add further depth to the analysis, we manually code 400 randomly selected sentences (200 pre-ChatGPT and 200 post-ChatGPT) that include at least one danger-related frame into the categories: 1) *Negative* (AI is a source of [potential] danger or risk), 2) *Neutral* (no clearly identifiable risk or presented as equally balanced with benefits) and 3) *Positive* (AI in a positive role, opposing the source of danger or harm). For both periods, the annotators' agreement was above 75%. We also annotated topical subgroups for each category (for more details on agreement and annotation, see Online Appendix 2). The most noticeable result of this analysis is the significant increase in the share of sentences where AI is the source of danger or risk, which is identified by both annotators (35% to 64% for A1 and 31% to 67.5% for A2). This strongly suggests that the overall narrative has shifted from seemingly balanced to much more alarmist. The shift appears to be driven by sentences related to application-specific dangers, such as AI-assisted fraud and abuse. While, for example, deepfakes have been a concern throughout the entire time period covered by the dataset, ChatGPT and other accessible text-generating tools have opened up entirely new possibilities—from cheating on school exams to sophisticated email fraud, and this shows in the qualitative review. A second driving factor is sentences that include very broad or unspecified threats from AI, which approximately doubles for both annotators (13% to 28% and 16% to 30.5% of danger-related frames respectively). Meanwhile, the variety of specific causes of harm also grows, including reports of negative mental health effects of interacting with chatbots.

All in all, our analysis suggests that if the dangers and risks of AI were a prevalent theme before ChatGPT, the launch pushed it into an even more central focus, both in terms of the number of articles mentioning it and the relative frequency with which it appears within these articles. This shift can largely be attributed to concerns around more sophisticated fraud schemes but also appears to be a result of a more general, unspecified worry around AI.

### *4.3.2 AI Anthropomorphism*

Perceptions of AI anthropomorphism, particularly in relation to language models, have recently received significant attention in the literature (Abercrombie et al., 2023; Cheng et al., 2024; Inie et al., 2024). To contribute to this ongoing discussion, we investigate which forms it has taken in the dominant news media that form public discourse and perceptions. As this is most appropriate given our tools, we consider *anthropomorphising language* rather than the broader notion of anthropomorphism itself. For example, our interest is not in whether authors themselves attribute human-like traits to AI but merely in that they are using language normally used to describe humans.

The concept of anthropomorphism is multifaceted and harder to link to frames than the concept of danger, so instead of relying on frames as a whole, we target only sentences where AI is part of the frame elements that indicate potential anthropomorphism. This illustrates the high specificity with which frame analysis can be conducted. This specificity is an advantage of relying on the FrameNet definitions but also presents a methodological limitation: by defining anthropomorphism through frames, we limit its scope to AI being presented as able to perform cognitive and communicative functions actively. Thus, anthropomorphising of other kinds, such as referring to chatbots using personal pronouns or assigning human-like names to AI products, are not explicitly considered here. Our manually curated list of potentially anthropomorphising frames and FEs pairs, therefore, focuses on AI taking roles such as *Cognizer* or *Speaker* (see Figure 6 for the full list of pairs).

To filter the sentences, we use the same regular expression as in Section 4.3: `(^.{0,3} |^)(AI).{0,3}$`. As it goes beyond the individual frame level with filtering, it is more useful to look at the frequencies of the sentences where such combinations of frames and FEs occur.

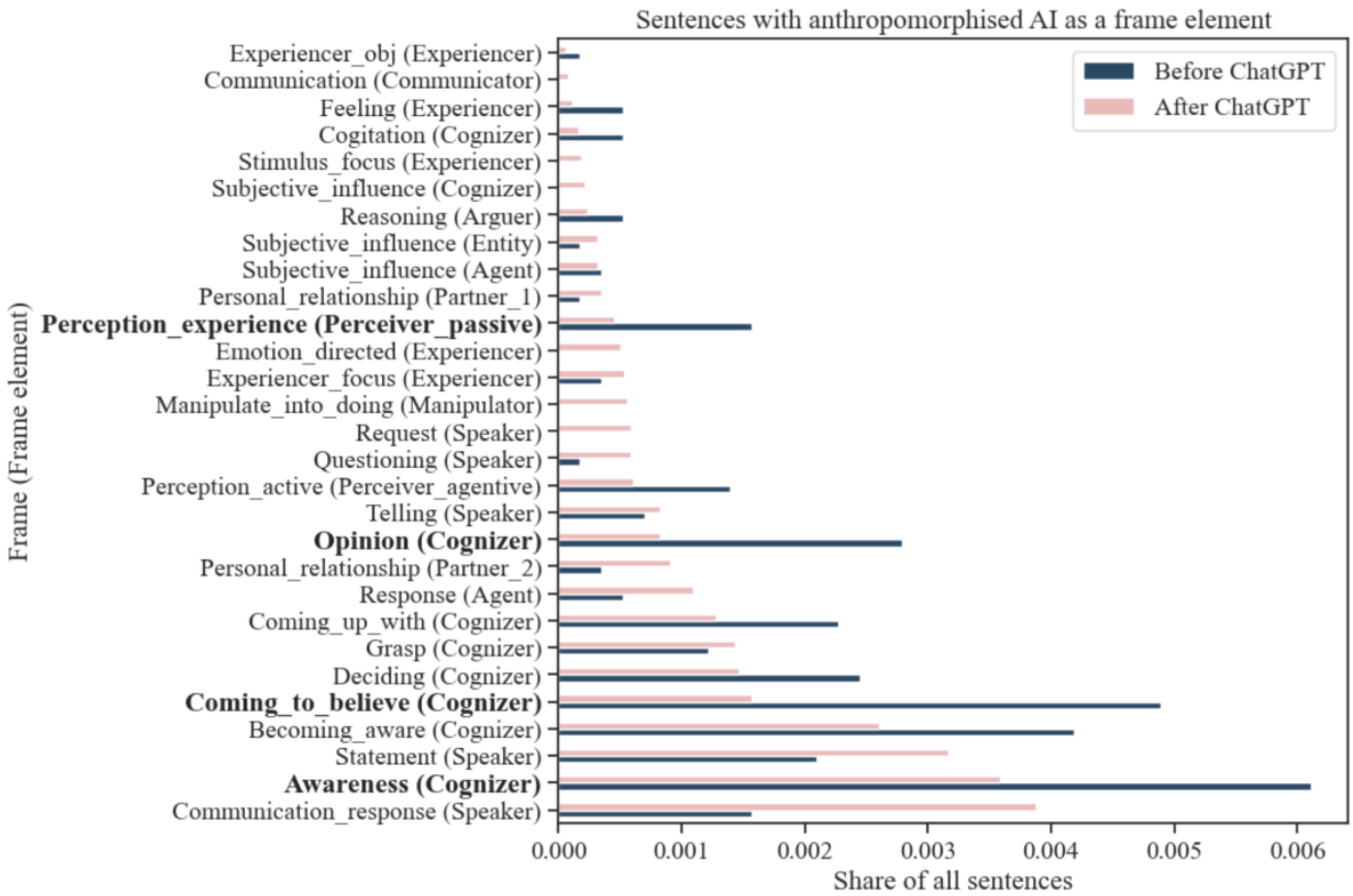

**Fig. 6** Frame-FE pairs related to anthropomorphism of AI before and after the release of ChatGPT. Statistically significant changes ($p<0.01$) in bold

Looking at the relative number of occurrences in more detail (Figure 6), we observe the changes in relative frequencies of sentences with frame-FE combinations. Due to the minute size of the subset of the sentences in the first half of the dataset, the shifts in relative frequencies are individually significant only for the most frequent pairs in the pre-ChatGPT half. Most of these significant changes happen in situations where AI assumes the role of the *Cognizer*. These frames, such as *Awareness*, *Coming_to_believe*, *Becoming_aware* are generally associated with cognitive functions, so they can be triggered both by direct implications of thought process ('AI thinks') and by casual and non-colloquial reference to technical terms such as machine learning ('Meta wants you to converse with it as the AI can learn from discussing things with humans.').

Due to the low number of sentences in the first six months, these changes do not appear statistically significant, but the two pairs with the highest relative growth in frequency (*Statement*, *Communication_response*) both notably cast AI as a *Speaker*. This can be straightforwardly explained by the emergence of chatbots. Almost always, when referring to a piece of text produced by the language model, it is referred to as something it 'said', 'claimed' or similar (e.g. '"To me, the soul is a concept of the animating force behind consciousness and life itself," the AI wrote.'). In total, we find that the first half of our sentence collection contains 190 (or 3.22%) relevant sentences, while the second half contains 1011 (2.33%), which constitutes a meaningful relative decrease ($p < 0.01$). A plausible interpretation would be that AI as a term became used in broader contexts, which is partially supported by our results in Section 4.2, showing a general increase of frames such as *Leadership* in our sentences. Additionally, as we show below, the first half of the period has already included novel anthropomorphising AI tropes unrelated to ChatGPT.

To complete the investigation, we move to the final step of our analysis. Similarly to the case of AI danger, we read and manually annotate a subsample of sentences, this time all 190 sentences from the first six months and 200 sentences sampled from the second six months. We divide them into four main groups: 1) *no anthropomorphism* (or inconclusive/unclear sentence), 2) *established anthropomorphising terms* (e.g. derivatives of academic and engineering terms ‘machine learning’ or ‘pattern recognition’), 3) *task-based anthropomorphism* (humanising language describing functionality) and 4) *high anthropomorphism* (implying cognitive abilities, feeling, human likeness, etc.). We further split the latter two into topical subgroups (see Online Appendix 2).

Unlike the coded danger-related sentences, the annotations of anthropomorphism-related sentences diverge more significantly between the two annotators, with agreement of only 65% and 59.5% before and after. We attribute this to the different professional backgrounds of the annotators: A1 has a computer science background and is much more likely to identify sentences as categories 1 and 2, while A2 has a background in the social sciences and tends to identify more complex anthropomorphism (categories 3 and 4). It does, however, hold for both annotators that the group that grew the most is task-specific anthropomorphism, and according to our subgroup annotations, this is largely due to the increase in the subgroup related to communication (see Online Appendix 2). This is almost certainly the effect that ChatGPT and other popular chatbots had: expressions such as ‘AI said’ or ‘we asked AI’ have become markedly more prevalent.

Our main observation concerning the subgroups was that more than 20% (42 and 48 out of 190 for A1 and A2 respectively) of the pre-ChatGPT sentences represent a specific news cliché about image generation applied to change the appearance of famous individuals, often based on age (‘Here's what AI thinks she'll look like at 50, 60, and 70.’). While very similar, such occurrences are unique and no doubt represent the attention to the novelty of AI tools for image generation released in the summer of 2022. When annotating the sentences, we choose to make this cliché a separate subgroup of high anthropomorphism. While such instances are still present after November 30, they constitute a considerably smaller share (11 and 17 out of 200, depending on the annotator). This subgroup accounts not only for the relative decrease of situations where AI plays the role of a *Cognizer* in frames *Awareness* and *Opinion*, but also possibly for the general relative decrease of anthropomorphising situations. This also highlights how difficult it is to disentangle the effects of one specific technology on the discourse. That is, our frame-based selection of anthropomorphising may have captured two separate effects. The dataset before the release of ChatGPT is dominated by the reporting on the recently-released image-generating models, while afterwards, it is displaced by ChatGPT’s impressive text generation.

There is reason to believe that the observations above are part of a larger ongoing process of anthropomorphising AI. Even with the methodological limitation to the detectable types of anthropomorphism, our manual subsample annotation highlights how anthropomorphising of the term stems from different sources, roughly corresponding to the subgroups we identified during the annotation process:

- Scientific and technical terms such as ‘pattern recognition’ or ‘machine learning’, inherently holding a degree of anthropomorphism, being adopted in common discourse and used non-idiomatically, which increases the effect;
- Anthropomorphising of convenience, to describe a system's functionality in non-technical and easily understandable language;
- Stylistic and narrative choices to attract readers or increase entertainment value;
- AI-related systems with social functions, such as AI-companions;
- Genuine projection of the capacity to think and feel onto the technology.

Finally, tightly related to AI anthropomorphism and some of the perceived dangers of the technology is the concept of Artificial General Intelligence (AGI), which can take a range of meanings but usually signifies an instance of AI that performs a wide range of human tasks as well as humans. The definition can sometimes extend to exhibiting (or even experiencing) consciousness or sentience or going far beyond human intelligence (‘superintelligence’). Naturally, there is no consensus on the exact definition of AGI, and it is unclear whether it is implementable. The term, however, has been used in the media to distinguish the futuristic theoretical superintelligence from the ubiquitous AI, now used to describe even simple single-purpose algorithms. Similarly, several AI companies, such as OpenAI, have claimed developing AGI as their mission, presenting it as a longer-term goal compared to their current projects (OpenAI, 2023). In our dataset, we find, somewhat unexpectedly, that the proportion of articles mentioning AGI in some form in the two halves of the timeframe is almost identical at ~1.9%: 15 articles before the release of ChatGPT and 100 articles after. Then again, sentient AI and existential threats have been historically featured in the AI discourse (Brennen et al., 2019; Fast and Horvitz, 2017), so even proportional growth may account for the increased visibility. It also remains possible that there are articles that mention AGI that never explicitly mention non-general artificial intelligence, which we have omitted while collecting our dataset.

# 5. Discussion

Our proposed methodology proves successful in achieving its main purpose: combining high-level quantitative analysis of a large corpus while making it significantly easier to identify themes that deserve additional attention through qualitative analysis. In addition to addressing the research questions about AI coverage, by encoding the sentences manually, we are able to confirm the validity of the annotation by LOME, as nearly all the sampled sentences can be seen as appropriate for containing respective frames. Yet, the method does come with disadvantages. Most notably, to limit the considerable processing time, we restricted ourselves to parsing only the sentences mentioning AI explicitly, leaving out the possibility of using coreference resolution to identify implicit mentions of the technology. The frames themselves only represent formalised semantic structures and do not translate into meanings or narrative elements directly; they serve as proxies in our methodology. The frequencies and occurrences of frames can be interpreted and related to the news coverage in the dataset but do not allow to establish causality.

This being said, our findings point to some unambiguous trends which can almost certainly be attributed to the release of ChatGPT. The most important and apparent is, of course, the massive increase in AI-related news coverage, which brought the topic to the attention of outlets that had previously given only marginal attention to it. All of our findings should be assessed in light of this quantitative increase. While we compare the number of articles or the presence of certain frames in *relative* terms, the increase in absolute values is such that it is safe to say that almost any AI-adjacent topic would take more space in the news coverage after November 2022 compared to before. In other words, when we say that, for example, the degree of anthropomorphic terms has decreased, this essentially means that a media consumer who reads *only* about AI is less likely to come across it. A general reader who consumes random news, on the other hand, is far more likely to encounter such tropes today than before November 2022, given that the likelihood of coming across an AI-related news article has drastically increased.

Moreover, the analysis of the most frequent semantic frames after the release date seems to suggest that more mentions of AI in relative terms belong to sentences with quotations and comments, which is in line with the general effort to demystify AI and LLMs identified by Roe and Perkins (2023). While we do not look at the sources of the statements directly, we also separately find a significant increase in mentions of government agencies and people in leadership positions (via the *Leadership* frame), which suggests that industrial leaders and governments play a greater role in AI discourse post-ChatGPT. In their 2018 paper, Brennen et al. raise concerns that the AI debate in the mass media is dominated by the industrial perspective, while voices from political activists, the general public and academia receive less attention. ChatGPT, being a commercial product, seems to have exacerbated this problem.

One of the key questions about the coverage of AI is how it is framed in terms of risks and benefits. However, the formulation and approach varies across studies, making comparisons challenging. Some studies aggregate metrics of emotional valence (e.g. sentiments such as positivity and negativity) over time. For instance, Fast and Horvitz (2017) observe that since 2010, the discussion has become much more optimistic. Garvey and Maskal (2020) apply sentiment analysis but do not find negative bias in either a single-source longitudinal dataset (NYT coverage between 1956 and 2018) or a broad one (NewsAPI coverage for three months in 2017–2018). On the other hand, Nguyen and Hekman (2022) apply sentiment analysis to a dataset of specialised and established media and note gradually more negative sentiment throughout the 2010s. Others focus on identifying what types of risks are prevalent in the coverage and how they are communicated (Brantner and Saurwein, 2021; Brennen et al., 2018; Köstler and Ossewaarde, 2022). Our study contributes to this debate by showing that the post-ChatGPT coverage shifted towards a more cautious and even alarmist stance. The relative proportion of the relevant frames grew significantly, suggesting dangers and risks taking a larger topic in the coverage. The manual coding of the sentences with danger-related frames further demonstrated how media attention shifted towards risks associated with AI, as opposed to AI being a solution for other problems or playing another positive role. The question now is whether this trend will last over time.

Another key contribution pertains to anthropomorphism. In their recent paper, Inie et al. (2024) provide a useful literature-based categorisation of the *classes* of anthropomorphising language: presenting a machine as a (1) cogniser, (2) agent, (3) biological system, or (4) communicator. Their methodology involves using FrameNet definitions to describe classes (1) and (4), which means that our definition closely maps to classes (1) and (4) as per Inie et al. (2024). In our work, and through coding in particular, we attempt to provide a different type of categorisation which focuses on the 'strength' and source of the anthropomorphising language. There is a much smaller leap in assigning human-like qualities when an already established anthropomorphising technical term is reused, than when cognitive abilities are projected onto a machine to provide a narrative. Different types of anthropomorphism can be tied to various tasks performed by autonomous systems, so we expect more diversity in anthropomorphising language as an increasing number of these systems are marketed as AI. Even within our year-long dataset, we find at least two recent tropes: the image-generation AI 'thinking' and chatbots being presented as human-like communicators. Finally, it is worth noting that a significant additional level of human likeness is clearly added when reporting the news (Cheng et al., 2024), though a large part of it is arguably inherited from the original research and industrial press releases.

While our focus is on how the release of ChatGPT affected AI coverage and the more granular changes in the use of the concept, it is worth reflecting on the notion of 'AI' itself. There is a possibility that, due to the popularity and marketing framing, AI has become at least temporarily synonymous with chatbots. When working on annotating sentences, we found that AI and ChatGPT are indeed often used interchangeably. However, comparing the most common frame roles for both terms beyond being an Instrument (Figure 4), we see considerable difference: AI takes a much more diverse set of roles, and the composition of these does not shift significantly. It would, however, be incorrect to dismiss chatbots and ChatGPT as having no bearing on the meaning of AI: Their sheer magnitude as a news event and the noticeable changes in coverage that followed are evidence to the contrary. Rather, we believe that the popularity of ChatGPT with the following media coverage brought a number of new meanings, functions and interpretations under the umbrella of AI, making it even broader, rather than taking it over. We can see this as a particularly prominent instance of AI—at least in the news media context—adopting a new set of meanings, essentially behaving as a floating signifier. Our various observations, including more common mentions of vague and unspecified AI risks and the growing range of common ways to anthropomorphise the technology, exemplify this process. Given the prominence of AI technologies, it is therefore crucial that we continue to study and scrutinise how the media construes the technological leaps that will come to define it. We invite other scholars to join us in this task and will happily share our data on request.

# References

Abercrombie, G., Curry, A. C., Dinkar, T., Rieser, V., & Talat, Z. (2023). *Mirages. On Anthropomorphism in Dialogue Systems*. 4776–4790.

Baker, C. F., Fillmore, C. J., & Lowe, J. B. (1998). The Berkeley FrameNet Project. *COLING 1998 Volume 1: The 17th International Conference on Computational Linguistics*. COLING 1998. https://aclanthology.org/C98-1013

Brantner, C., & Saurwein, F. (2021). Covering Technology Risks and Responsibility: Automation, Artificial Intelligence, Robotics, and Algorithms in the Media. *International Journal of Communication*, *15*, 5074–5098.

Brennen, J. S., Howard, P. N., & Nielsen, R. K. (2018). *An Industry-Led Debate: How UK Media Cover Artificial Intelligence*.

Brennen, J., Schulz, A., Howard, P., & Nielsen, R. (2019). *Industry, Experts, or Industry Experts? Academic Sourcing in News Coverage of AI*.

Cave, S., Craig, C., Dihal, K., Dillon, S., Montgomery, J., Singler, B., & Taylor, L. (2018). *Portrayals and perceptions of AI and why they matter*.

Cheng, M., Gligoric, K., Piccardi, T., & Jurafsky, D. (2024, March). AnthroScore: A Computational Linguistic Measure of Anthropomorphism. *Proceedings of the 18th Conference of the European Chapter of the Association for Computational Linguistics*.

Chuan, C.-H., Tsai, W.-H. S., & Cho, S. Y. (2019). Framing Artificial Intelligence in American Newspapers. *Proceedings of the 2019 AAAI/ACM Conference on AI, Ethics, and Society*, 339–344. https://doi.org/10.1145/3306618.3314285

Fast, E., & Horvitz, E. (2017). *Long-term trends in the public perception of artificial intelligence*. *31*(1).

Frost, E. K., & Carter, S. M. (2020). Reporting of screening and diagnostic AI rarely acknowledges ethical, legal, and social implications: A mass media frame analysis. *BMC Medical Informatics and Decision Making*, *20*(1), 325. https://doi.org/10.1186/s12911-020-01353-1

Garvey, C., & Maskal, C. (2020). Sentiment Analysis of the News Media on Artificial Intelligence Does Not Support Claims of Negative Bias Against Artificial Intelligence. *OMICS: A Journal of Integrative Biology*, *24*(5), 286–299. https://doi.org/10.1089/omi.2019.0078

Inie, N., Druga, S., Zukerman, P., & Bender, E. M. (2024). *From 'AI' to Probabilistic Automation: How Does Anthropomorphization of Technical Systems Descriptions Influence Trust?* 2322–2347.

Köstler, L., & Ossewaarde, R. (2022). The making of AI society: AI futures frames in German political and media discourses. *AI & SOCIETY*, *37*(1), 249–263. https://doi.org/10.1007/s00146-021-01161-9

Lupton, D. (2021). 'Flawed', 'Cruel' and 'Irresponsible': The Framing of Automated Decision-Making Technologies in the Australian Press. *SSRN Electronic Journal*. https://doi.org/10.2139/ssrn.3828952

Majid, A. (2024, April 15). Top 50 biggest news websites in the world: Newsweek doubles visits year-on-year in March. *Press Gazette*. https://pressgazette.co.uk/media-audience-and-business-data/media_metrics/most-popular-websites-news-world-monthly-2/

Nader, K., Toprac, P., Scott, S., & Baker, S. (2022). Public understanding of artificial intelligence through entertainment media. *AI & SOCIETY*. https://doi.org/10.1007/s00146-022-01427-w

Nguyen, D., & Hekman, E. (2022). The news framing of artificial intelligence: A critical exploration of how media discourses make sense of automation. *AI & SOCIETY*. https://doi.org/10.1007/s00146-022-01511-1

Ouchchy, L., Coin, A., & Dubljević, V. (2020). AI in the headlines: The portrayal of the ethical issues of artificial intelligence in the media. *AI & SOCIETY*, *35*(4), 927–936. https://doi.org/10.1007/s00146-020-00965-5

*Planning for AGI and beyond*. (n.d.). Retrieved 27 May 2024, from https://openai.com/index/planning-for-agi-and-beyond/

Roe, J., & Perkins, M. (2023). 'What they're not telling you about ChatGPT': Exploring the discourse of AI in UK news media headlines. *Humanities and Social Sciences Communications*, *10*(1), 1–9. https://doi.org/10.1057/s41599-023-02282-w

Sun, S., Zhai, Y., Shen, B., & Chen, Y. (2020). Newspaper coverage of artificial intelligence: A perspective of emerging technologies. *Telematics and Informatics*, *53*, 101433. https://doi.org/10.1016/j.tele.2020.101433

Xia, P., Qin, G., Vashishtha, S., Chen, Y., Chen, T., May, C., Harman, C., Rawlins, K., White, A. S., & Van Durme, B. (2021). LOME: Large Ontology Multilingual Extraction. *EACL 2021*, 149.

Zhai, Y., Yan, J., Zhang, H., & Lu, W. (2020). Tracing the evolution of AI: Conceptualization of artificial intelligence in mass media discourse. *Information Discovery and Delivery*, *48*(3), 137–149. https://doi.org/10.1108/IDD-01-2020-0007

# Online Appendix 1: Significance testing for frame analysis

**Table 1** Significance tests for the differences in the number of occurrences before and after the release of ChatGPT for the most common frames (Figure 3b). Contains top-15 frames from both halves of the dataset (17 in total). p-values<0.01 in bold.

| Frame | Occurrences before | Occurrences after | Chi2 | p-value |
|---|---|---|---|---|
| *Artifact* | 574 | 4546 | 0.9525 | 0.32909 |
| *Awareness* | 566 | 3886 | 4.7853 | 0.02870 |
| ***Businesses*** | **895** | **5901** | **15.2241** | **<1e-05** |
| ***Calendric_unit*** | **521** | **4816** | **18.6392** | **<1e-05** |
| *Capability* | 1103 | 8296 | 0.0515 | 0.82047 |
| *Creating* | 502 | 4116 | 2.7191 | 0.09915 |
| ***Education_teaching*** | **599** | **5099** | **7.2197** | **0.00721** |
| *Gizmo* | 920 | 7030 | 0.0468 | 0.82880 |
| ***Increment*** | **474** | **4233** | **11.4674** | **0.00071** |
| ***Information*** | **553** | **3022** | **50.5329** | **<1e-05** |
| ***Leadership*** | **303** | **3944** | **84.7286** | **<1e-05** |
| *People* | 705 | 5421 | 0.117 | 0.73231 |
| ***Physical_artworks*** | **527** | **2714** | **66.8061** | **<1e-05** |
| ***Quantified_mass*** | **739** | **4911** | **11.1374** | **0.00085** |
| ***Statement*** | **1242** | **13,004** | **123.2716** | **<1e-05** |
| ***Text*** | **516** | **6115** | **97.8167** | **<1e-05** |
| *Using* | 1287 | 9402 | 1.5384 | 0.21486 |
| **All frames** | 48,972 | 371,183 | | |

**Table 2** Significance tests for the differences in the number of occurrences before and after the release of ChatGPT for the most common frames-FE pairs where AI is the frame element (Figure 4). Contains top-10 frames-FE pairs from both halves of the dataset (12 in total). p-values<0.01 in bold.

| Frame (Frame element) | Occurrences before | Occurrences after | Chi2 | p-value |
|---|---|---|---|---|
| *Artifact* (*Type*) | 154 | 980 | 0.9361 | 0.33328 |
| *Assistance* (*Helper*) | 67 | 455 | 0.0094 | 0.92256 |
| *Businesses* (*Product*) | 40 | 312 | 0.3865 | 0.53412 |
| *Capability* (*Entity*) | 202 | 1318 | 0.6252 | 0.42913 |
| *Creating* (*Creator*) | 47 | 238 | 3.606 | 0.05757 |
| ***Gizmo*** **(*Use*)** | **271** | **2403** | **15.8234** | **<1e-05** |
| *Intentionally_act* (*Agent*) | 45 | 340 | 0.2159 | 0.64216 |
| ***Objective_influence*** **(*Influencing_entity*)** | **23** | **284** | **6.881** | **0.00871** |
| *Progression* (*Entity*) | 60 | 578 | 5.7397 | 0.01659 |
| ***Project*** **(*Salient_entity*)** | **53** | **170** | **24.4175** | **<1e-05** |
| *Research* (*Topic*) | 27 | 297 | 5.0241 | 0.025 |
| *Using* (*Instrument*) | 373 | 2336 | 3.5538 | 0.05941 |
| **All pairs** | 4119 | 28,573 | | |

**Table 3** Significance tests for the differences in the number of occurrences before and after the release of ChatGPT for the frames related to danger (Figure 5). p-values<0.01 in bold.

| Frame | Occurrences before | Occurrences after | Chi2 | p-value |
|---|---|---|---|---|
| *Arrest* | 6 | 100 | 3.1416 | 0.07632 |
| *Attack* | 53 | 344 | 0.9494 | 0.32988 |
| *Bearing_arms* | 7 | 82 | 0.9012 | 0.34246 |
| ***Being_at_risk*** | **110** | **1156** | **10.5681** | **0.00115** |
| *Catastrophe* | 65 | 442 | 0.5604 | 0.4541 |
| *Cause_harm* | 13 | 79 | 0.3333 | 0.56372 |
| *Defending* | 23 | 98 | 5.6602 | 0.01735 |
| *Destroying* | 25 | 212 | 0.185 | 0.66714 |
| *Fear* | 13 | 156 | 2.2085 | 0.13725 |
| *Hostile_encounter* | 60 | 596 | 3.7775 | 0.05195 |
| *Killing* | 45 | 430 | 1.9918 | 0.15816 |
| *Military* | 28 | 148 | 2.6941 | 0.10072 |
| ***Predicament*** | **98** | **452** | **19.7159** | **<1e-05** |
| *Prevarication* | 11 | 127 | 1.4798 | 0.22381 |
| ***Risky_situation*** | **116** | **1644** | **43.5367** | **<1e-05** |
| *Run_risk* | 75 | 586 | 0.0351 | 0.85141 |
| *Theft* | 21 | 224 | 1.9749 | 0.15993 |
| ***Warning*** | **37** | **845** | **47.054** | **<1e-05** |
| *Weapon* | 55 | 425 | 0.0041 | 0.94923 |
| **All frames** | 48,972 | 371,183 | | |

**Table 4** Significance test for the differences in the number of sentences including the anthropomorphising frame-FE pairs before and after the release of ChatGPT (Figure 6). p-values<0.001 in bold.

| Frame (Frame element) | Sentences before | Sentences after | Chi2 | p-value |
|---|---|---|---|---|
| ***Awareness* (*Cognizer*)** | **35** | **135** | **12.3262** | **0.00045** |
| *Becoming_aware* (*Cognizer*) | 24 | 98 | 6.8501 | 0.00886 |
| *Cogitation* (*Cognizer*) | 3 | 6 | 2.2679 | 0.13208 |
| ***Coming_to_believe* (*Cognizer*)** | **28** | **59** | **33.6419** | **<1e-05** |
| *Coming_up_with* (*Cognizer*) | 13 | 48 | 4.6176 | 0.03165 |
| *Communication* (*Communicator*) | 0 | 3 | 0 | 1.0 |
| *Communication_response* (*Speaker*) | 9 | 146 | 4.625 | 0.03151 |
| *Deciding* (*Cognizer*) | 14 | 55 | 4.1848 | 0.04079 |
| *Emotion_directed* (*Experiencer*) | 0 | 19 | 1.5053 | 0.21985 |
| *Experiencer_focus* (*Experiencer*) | 2 | 20 | 0.0019 | 0.96483 |
| *Experiencer_obj* (*Experiencer*) | 1 | 2 | 0.0728 | 0.78726 |
| *Feeling* (*Experiencer*) | 3 | 4 | 3.9294 | 0.04745 |
| *Grasp* (*Cognizer*) | 7 | 54 | 0 | 1.0 |
| *Manipulate_into_doing* (*Manipulator*) | 0 | 21 | 1.7574 | 0.18494 |
| ***Opinion* (*Cognizer*)** | **16** | **31** | **20.7392** | **<1e-05** |
| *Perception_active* (*Perceiver_agentive*) | 8 | 23 | 4.7249 | 0.02973 |
| ***Perception_experience* (*Perceiver_passive*)** | **9** | **17** | **11.1617** | **0.00084** |
| *Personal_relationship* (*Partner_1*) | 1 | 13 | 0.0123 | 0.91172 |
| *Personal_relationship* (*Partner_2*) | 2 | 34 | 0.7792 | 0.37739 |
| *Questioning* (*Speaker*) | 1 | 22 | 0.5908 | 0.44211 |
| *Reasoning* (*Arguer*) | 3 | 9 | 0.9791 | 0.32243 |
| *Request* (*Speaker*) | 0 | 22 | 1.8843 | 0.16984 |
| *Response* (*Agent*) | 3 | 41 | 0.5885 | 0.443 |
| *Statement* (*Speaker*) | 12 | 119 | 0.5747 | 0.4484 |
| *Stimulus_focus* (*Experiencer*) | 0 | 7 | 0.139 | 0.70932 |
| *Subjective_influence* (*Cognizer*) | 0 | 8 | 0.2277 | 0.63323 |
| *Subjective_influence* (*Agent*) | 2 | 12 | 0 | 1.0 |
| *Subjective_influence* (*Entity*) | 1 | 12 | 0.0002 | 0.98863 |
| *Telling* (*Speaker*) | 4 | 31 | 0 | 1.0 |
| **All sentences** | 5730 | 49120 | | |

# Online Appendix 2: Annotation process

Here, we present the more detailed results of the thematic coding we performed in our study.
For both narrow topics considered in the paper—AI danger and AI anthropomorphism—we applied the same procedure. Two annotators—A1 (computer science background) and A2 (social sciences background)—independently read and annotated 200 sentences from the articles before and after the release of ChatGPT for both topics (with the exception of anthropomorphising sentences pre-ChatGPT, all 190 of them were annotated). The sentences themselves are selected based on the presence of certain semantic frames in them (see Sections 4.3.1–4.3.2). To annotate danger-related sentences, as the first step, both annotators coded the sentences using three categories: 1) *Negative* (AI is a source of [potential] danger or risk), 2) *Neutral* (no clearly identifiable risk and benefits presented as equal) and 3) *Positive* (AI in a positive role, opposing the source of danger or harm). At this level, the annotators' agreement is 78.5% before ChatGPT (Cohen's kappa $\kappa$=0.675) and 81.5% after ($\kappa$=0.637). For the second level of annotations, both annotators attempted to find distinct patterns and subcategories for each of the main categories. The breakdown and short description of each subcategory are presented in Table 5.

**Table 5** Thematic groupings of the sentences related to danger in the random subsamples from before and after the release of ChatGPT.

| First level annotation | Second level annotation | Before ChatGPT | | After ChatGPT | |
|---|---|---|---|---|---|
| | | **A1** | **A2** | **A1** | **A2** |
| AI as a source of danger | Existential risks: danger to humanity, extinction, etc. | 23 | 21 | 33 | 35 |
| | Job market-related risks | 6 | 5 | 11 | 16 |
| | Criminal and illegal activities, cheating | 9 | 4 | 17 | 16 |
| | Specific reported cases | 6 | 0 | 11 | 7 |
| | Unspecified or broad risks and dangers | 26 | 32 | 56 | 61 |
| | **Negative** | **70** | **62** | **128** | **135** |
| Neutral/ inconclusive | Inconclusive or balanced | 44 | 68 | 33 | 32 |
| | Danger is part of a fictional reference or prompting output | 16 | 0 | 1 | 0 |
| | Military AI coverage with unclear positioning | 14 | 12 | 3 | 4 |
| | **Neutral** | **74** | **80** | **37** | **36** |
| AI in a positive role | AI as a solution: AI applications that mitigate risks | 54 | 57 | 30 | 24 |
| | Ethical AI: mentions of 'good', 'ethical' or 'fair' AI as opposed to 'normal' AI | 2 | 0 | 4 | 3 |
| | AI as an entity exposed to danger | 0 | 1 | 1 | 2 |
| | **Positive** | **56** | **58** | **35** | **29** |
| **Annotated in total** | | 200 | | 200 | |

The annotation of the anthropomorphism-related sentences was done in a similar way: first, both annotators coded the sentences with four categories: 1) *no anthropomorphism* (or inconclusive/unclear sentence), 2) *established anthropomorphising terms* (e.g. derivatives of academic and engineering terms 'machine learning' or 'pattern recognition'), 3) *task-based anthropomorphism* (humanising language describing functionality) and 4) *high anthropomorphism* (implying cognitive abilities, feeling, human likeness, etc.). The annotators' agreement for individual annotations was lower: 65% for the sentences from before the release of ChatGPT ($\kappa$=0.545) and 59.5% for the sentences from after the event ($\kappa$=0.425). As we briefly discuss in Section 4.3.2, we attribute a lower level of agreement, at least partially, to the different professional backgrounds of the annotators. As the second step, the annotators attempted to split categories 3 and 4 into smaller subcategories representing more specific types of anthropomorphism or media tropes. The breakdown and short description of each subcategory are presented in Table 6.

**Table 6** Thematic groupings of the sentences related to anthropomorphism in the random subsamples from before and after the release of ChatGPT.

| First level annotation | Second level annotation | Before ChatGPT | | After ChatGPT | |
|---|---|---|---|---|---|
| | | **A1** | **A2** | **A1** | **A2** |
| No anthropomorphism or unclear sentence | | **38** | **16** | **25** | **35** |
| Established anthropomorphising terms (e.g. machine learning) | | **27** | **19** | **39** | **11** |
| Task-based anthropomorphism (explaining functions in 'human' terms) | Composing, creating | 10 | 31 | 14 | 32 |
| | Communicating and producing natural language | 11 | 19 | 40 | 53 |
| | Collecting information | 13 | 19 | 9 | 4 |
| | Making a choice or a decision | 12 | 2 | 9 | 2 |
| | **Total task-based** | **46** | **71** | **72** | **91** |
| High anthropomorphism | AI is described as having cognitive abilities beyond describing its formal tasks (thinking, curiosity, etc.), excluding the case below | 23 | 27 | 23 | 34 |
| | The specific news cliche about image generation: e.g. 'Here is how AI thinks X would look if…' | 42 | 48 | 11 | 17 |
| | AI presented as having emotions, or taking emotionally charged actions | 12 | 7 | 21 | 2 |
| | AI taking human roles | 2 | 2 | 9 | 10 |
| | **Total high-level** | **79** | **84** | **64** | **63** |
| **Annotated in total** | | 190 | | 200 | |